\documentclass[10pt,twocolumn,letterpaper]{article}

\usepackage[pagenumbers]{cvpr} % To force page numbers, e.g. for an arXiv version

\usepackage{amsmath}
\usepackage[dvipsnames]{xcolor}

\newcommand{\norm}[1]{\left\lVert#1\right\rVert}
\definecolor{cvprblue}{rgb}{0.21,0.49,0.74}

\definecolor{myred}{rgb}{1.0, 0.6, 0.5}
\definecolor{mygreen}{rgb}{0.39, 0.74, 0.21}
\definecolor{myyellow}{rgb}{1.0, 0.9, 0.4}
\definecolor{myblue}{rgb}{0.11,0.29,0.54}

\usepackage[pagebackref,breaklinks,colorlinks,citecolor=cvprblue]{hyperref}
\newcommand\blfootnote[1]{%
  \begingroup
  \renewcommand\thefootnote{}\footnote{#1}%
  \addtocounter{footnote}{-1}%
  \endgroup
}

\newcommand\illumination{
    [~\includegraphics[width=.8em]{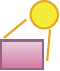}~]
}
\newcommand\wing{
    [~\includegraphics[width=.8em]{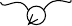}~]
}
\newcommand\orientation{
    [~\includegraphics[width=.8em]{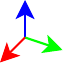}~]
}
\newcommand\dollyzoom{
    [~\includegraphics[width=.8em]{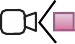}~]
}

\usepackage{amsmath,amsfonts,bm}

\usepackage{bbm}
\usepackage{bm}

\usepackage{booktabs}       % professional-quality tables
\usepackage{amsfonts}       % blackboard math symbols
\usepackage{nicefrac}       % compact symbols for 1/2, etc.
\usepackage{microtype}      % microtypography
\usepackage{color}
\usepackage{array}
\usepackage{multirow}
\usepackage{dirtytalk}
\usepackage{amsthm}
\usepackage[para,online,flushleft]{threeparttable}
\usepackage{colortbl}
\usepackage[normalem]{ulem}

\definecolor{Crimson}{rgb}{0.86, 0.08, 0.24}
\definecolor{DarkGreen}{rgb}{0.10, 0.55, 0.10}
\definecolor{RoyalBlue}{rgb}{0.20, 0.60, 0.86}
\definecolor{DarkCyan}{rgb}{0.0, 0.54, 0.54}
\definecolor{Gray}{gray}{0.9}
\definecolor{ChromeYellow}{rgb}{1.0, 0.65, 0.0}
\definecolor{Salmon}{rgb}{0.98, 0.50, 0.45}
\ifx \submission \undefined

\else
\fi

\begin{document}

%%%%%%%%% TITLE - PLEASE UPDATE Learning Continuous 3D Words
\title{ Learning Continuous 3D Words for Text-to-Image Generation}

\author{Ta-Ying Cheng$^{1}\footnotemark$ \ \ Matheus Gadelha$^{2}$ \ \ Thibault Groueix$^2$ \ \ Matthew Fisher$^2$ \\ \ \ Radom\'{i}r M\u{e}ch$^2$ \ \ Andrew Markham$^1$ \ \ Niki Trigoni$^1$
\vspace{0.2cm}
\\\hspace{-0.1cm}$^1$University of Oxford \ \ $^2$Adobe Research
\vspace{0.08cm}
\\
}

%%%%%%%%% AUTHORS - PLEASE UPDATE
% \and

% \input{author-kit-CVPR2024-v2/editing_env}
% \begin{document}
%\maketitle

% \twocolumn[{%
% \renewcommand\twocolumn[1][]{#1}%
% \maketitle

% \begin{center}
%     \centering
%     \includegraphics[width=\linewidth]{author-kit-CVPR2024-v2/figures/fig1_teaser.pdf}
%     \vspace{-7pt}
% \captionof{figure}{Given a few images of a new concept, our method augments a pre-trained text-to-image diffusion model, enabling generating complex compositions of the concept. Our method benefits existing methods (e.g., Custom Diffusion, Dreambooth.)}
%     \label{fig:teaser}
% \end{center}

% }]
% \maketitle
\twocolumn[{%
\renewcommand\twocolumn[1][]{#1}%
\maketitle
\begin{center}
    \centering
    \includegraphics[width=\linewidth]{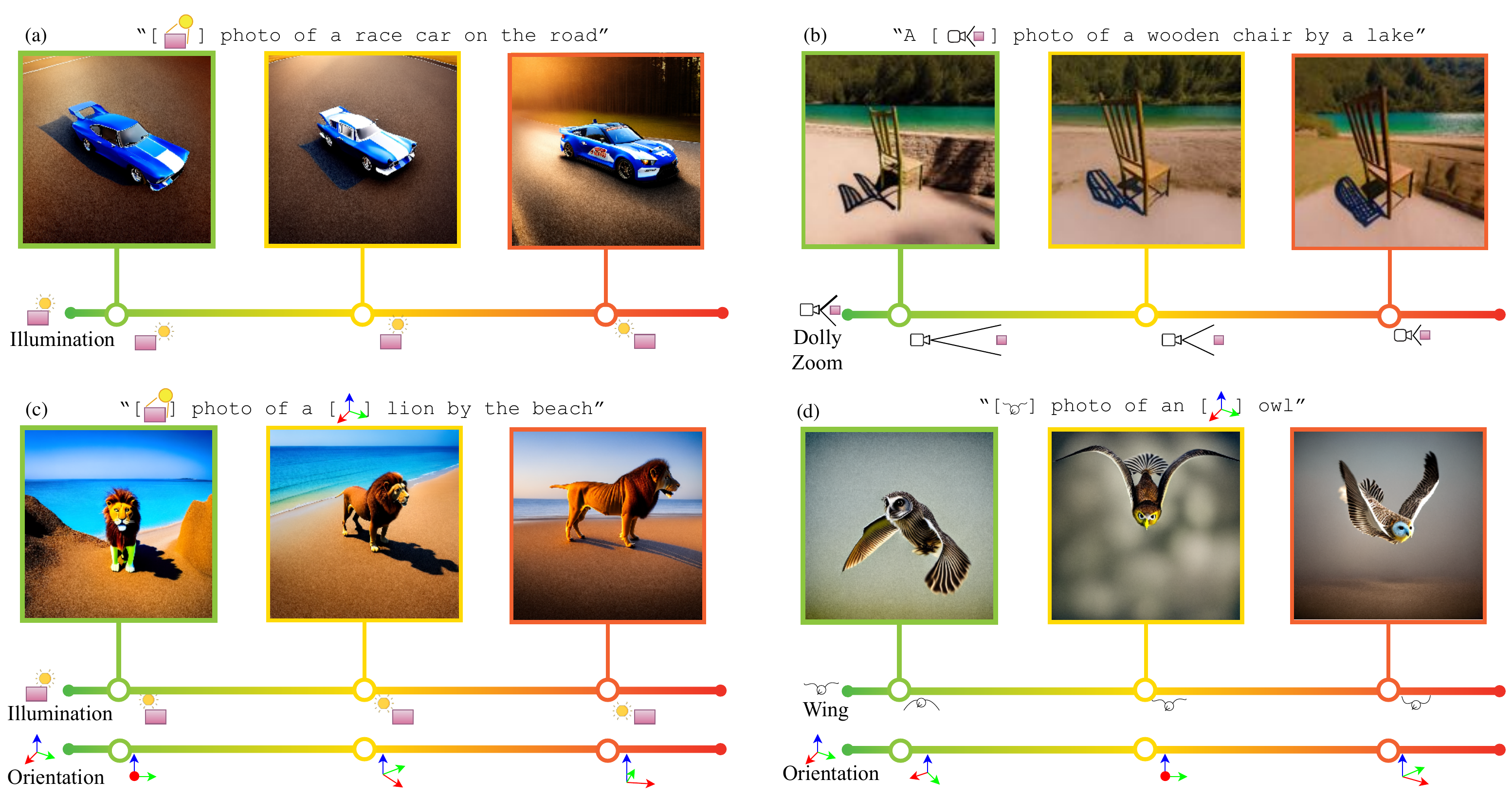}
    \vspace{-7pt}
\captionof{figure}{We introduce \textbf{Continuous 3D Words} -- special tokens in text-to-image models that allow users
to have fine-grained control over several attributes like illumination~\illumination (a and c), non-rigid shape change~\wing (d), orientation~\orientation (c and d), and camera parameters~\dollyzoom (b).
Our approach can be trained using a \emph{single 3D mesh} and a rendering engine while incurring into negligible 
runtime and memory costs.
}
    \label{fig:teaser}
\end{center}

}]

\begin{abstract}
\vspace{-3mm}
Current controls over diffusion models (e.g., through text or ControlNet) for image generation fall short in recognizing abstract, continuous attributes like illumination direction or non-rigid shape change. 
In this paper, we present an approach for allowing users of text-to-image models to have fine-grained 
control of several attributes in an image.
We do this by engineering special sets of input tokens that can be transformed in a continuous
manner -- we call them \textbf{Continuous 3D Words}.
These attributes can, for example, be represented as sliders and applied jointly with text prompts for fine-grained control over image generation. 
Given only a single mesh and a rendering engine, we show that our approach can be adopted to provide continuous user control over several 3D-aware attributes, including time-of-day illumination, bird wing orientation, dollyzoom effect, and object poses.
Our method is capable of conditioning image creation with multiple Continuous 3D Words and text descriptions simultaneously while adding no overhead to the generative process. Project Page: \href{https://ttchengab.github.io/continuous_3d_words}{https://ttchengab.github.io/continuous\_3d\_words}
\end{abstract}

\vspace{-5mm}    

\blfootnote{\textsuperscript{*}Work was done during internship at Adobe Research.}
\section{Introduction}
Photography is fascinating because it enables very detailed control over the composition and aesthetics of the final image. On the one hand, this is simply the result of application of physical laws to achieve image acquisition. On the other, the slightest changes in the moment captured, illumination, object orientation, or camera parameters bring a completely different feeling to the viewer.
While the giant leap of modern text-to-image diffusion can bring generated 2-D images to close proximity with real photos, text prompts are inherently limited to high-level descriptions, far removed from the detailed controls one has over actual photography. This is mainly due to the scarcity of such descriptions in the training dataset --- very few would describe a photo based on exact object movements and camera parameters like the wing pose of a bird or the rotation of a person's head in degrees. 
% Don't think we really need to talk about controlnet here
%While recent work such as ControlNet~\cite{controlnet} introduced additional fine-grained controls with image conditions, but extracting the aforementioned attributes from thees conditioned images and apply it to different objects remains an open challenge.
On the other hand, 3D rendering engines allow us to mimic many of these 3D controls that photographers enjoy. 
We can render images of objects with predefined camera, illumination and pose changes at a very fine-grained scale. 
However, creating detailed 3D worlds is incredibly laborious, which limits the diversity of the scenes 
that can be generated by non-specialized practitioners.
In that regard, using text-to-image diffusion to create images is a much more accessible technology, whereas precise 3D scene control remains firmly in the domain of experts.

In this work, we aim to bring together the best of two worlds by expanding the vocabulary of text-to-image diffusion models with very few samples generated from rendering engines. 
Specifically, we render meshes based on the attribute we aim to control, creating images with color and other useful information to generate a small set of data samples. 
The goal is to disentangle these abstract attributes from the original object and encode them into the textual space in a controllable manner -- we term these attributes \emph{Continuous 3D Words}.
They allow users to create custom sliders that enable fine-grained control during image generation and can be seamlessly used along text prompts.

At the heart of our approach is an algorithm to learn a continuous vocabulary. 
% MATHEUS: We kind of made the MLP before we knew about NETI, right? Not sure if mentioning it here helps the story.
% Inspired by recent work NETI which discovers the richness of a simple neural mapper in enhancing learning object concepts [CITE NETI], we use an MLP that takes the input of the attribute we aim to control and outputs a Continuous Word (feature token). 
The benefits of continuity are two-fold: \textit{i)} the association between different values of the same attribute makes it much easier to learn, rather than having to learn hundreds of discrete tokens as an approximation and \textit{ii)} we learn an MLP that would allow interpolation during inference to generate an actual continuous control.
On top of this, we also propose two training strategies to prevent degenerate solutions and enable generalization on new objects beyond the training category.
% \red{increase the editability of our image generation: unclear}. 
First, we apply a two-stage training strategy: we first apply the Dreambooth~\cite{ruiz2023dreambooth} approach to learn the object identity of the underlying mesh used for rendering, then sequentially learn the various attribute values disentangled from the object identity.
% associate the object with the same identity different attribute values, then try to learn their specific attributes values: unclear}. 
This prevents the model from falling into a degenerate solution of encoding each value of an attribute as a new object, which would prevent us from generalizing the attribute to new objects. Second, we apply ControlNet~\cite{controlnet} with various conditioned images to generate a set of additional images with varying backgrounds and object textures. This prevents the model from overfitting to the artificial backgrounds of rendered images. The entire training was done in a light-weight Lower Rank Adaptation (LoRA)~\cite{lora} manner, making it fast and accessible with single GPUs.

We implement our continuous vocabulary and training method, across various sets of single (e.g., dollyzoom extracted from chairs) and multiple (e.g., object pose and illumination extracted from a dog mesh) attributes, and show through quantitative user studies and qualitative comparisons that our method can properly reflect various attributes while maintaining the aesthetics of the image --- significantly outperforming competitive baselines.

In summary, we present 1) Continuous 3D Words, a new method of gaining 3D-aware, continuous attribute controls over text-to-image generation that can be easily tailored to a plethora of new conditions, 2) a series of training strategies to disentangle the attributes from object identity to enhance the improvements in image generation and 3) extensive qualitative and quantitative studies to showcase our approach in various interesting applications.

\section{Related Work}
\begin{figure*}[t]
\centering
\includegraphics[width=\linewidth]{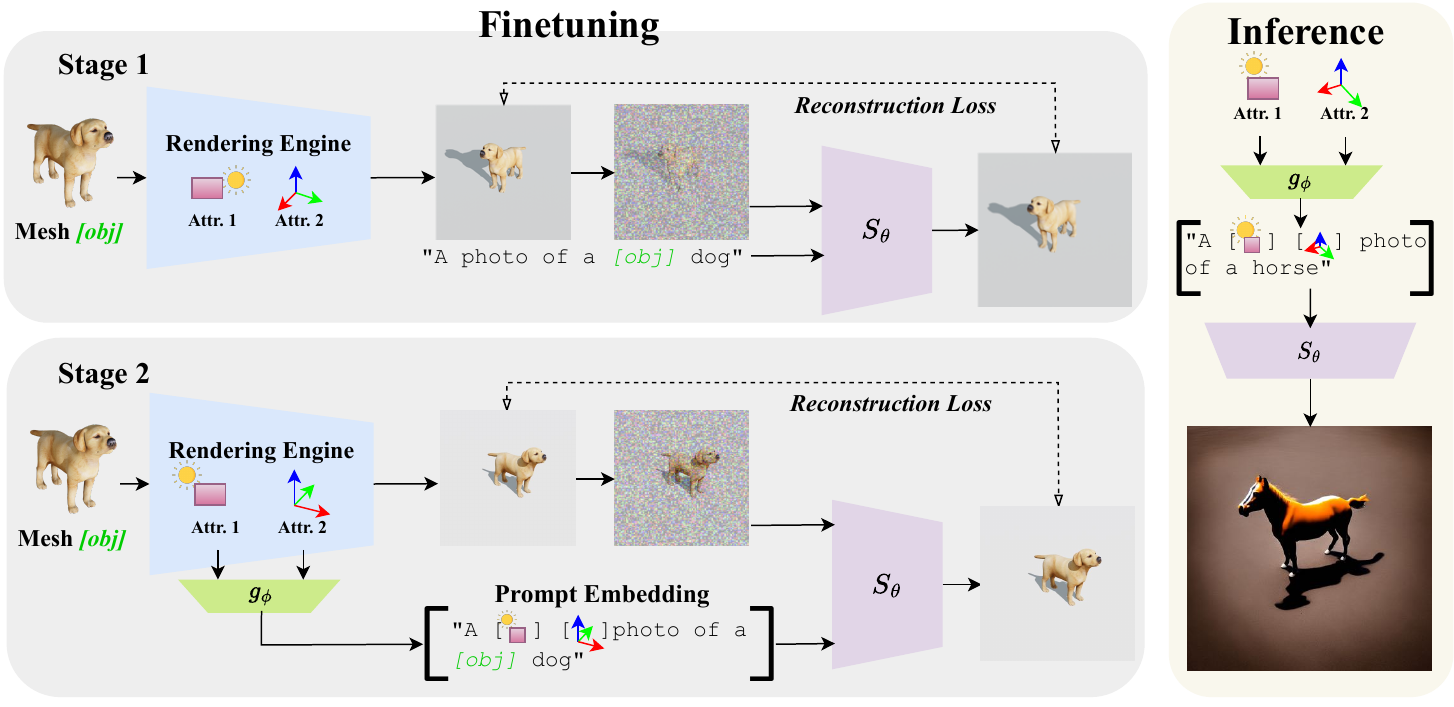}
% \vspace{-6mm}
\caption{
\label{fig:pipeline}
\textbf{Method Overview.} \textit{Finetuning:} Our finetuning is divided into two stages. In the first stage, we render a series of images using different attribute values (e.g., illumination and pose). We feed them into the text-to-image diffusion model to learn token embedding \textcolor{green}{\textit{[Obj]}} representing the \emph{single mesh} used for training. In the second stage, we add the tokens representing individual attributes into the prompt embedding. The two stage training allows us to better disentangle the individual attributes against \textcolor{green}{\textit{[Obj]}}. \textit{Inference:} Attributes can be applied to different objects for text-to-image generation.
}
% \vspace{-1mm}
\end{figure*}
\paragraph{Conditional Diffusion Models.}

Ever since diffusion models~\cite{ddpm, ddim} pushed the quality and generalizability of image generation beyond GANs~\cite{goodfellowgan}, the vision community has introduced a diverse range of modalities that can be used as conditions to control the image generation process.
The most common condition is currently text. 
Works such as DALLE~\cite{dalle1,dalle2,dalle3} and Imagen~\cite{imagen} used large scale text-image datasets and strong language understandings from pretrained LLMs~\cite{gpt3,t5, bert} to guide the generation process. 
Stable Diffusion further popularized this class of methods by employing memory efficient models
through latent-space diffusion~\cite{stablediffusion}. 
Other works built on top of these models by adding other forms of conditioning~\cite{controlnet,t2i}. 
Highly relevant to our work is ControlNet~\cite{controlnet}, which proposes a general pipeline with zero-convolutions for conditioning on text and image data (\eg., depth maps, canny maps, sketches).
Despite their impressive image quality, it is not clear how to use these models to control other
attributes of images like illumination or object orientation.

Other set of works explored how to perform image edits using textual instructions.
Given a text-generated image, they demonstrate how the user can edit the image by amending the prompt, yet still preserve some aspects of the original image~\cite{pix2pixzero, prompt2prompt,instructpix2pix}. 
While convenient, these approaches do not allow \emph{fine-grained} control over image elements since they are ultimately
restricted by the user's ability to describe visual content through text -- e.g., it would be very difficult to change
the illumination direction by a precise angle such as $11^{\circ}$.

Recently, as the amount of 3D data available significantly increased, Liu et al.~\cite{liu2023zero1to3} introduced Zero-1-to-3, a diffusion model trained on various viewpoints of 3D rendering that enables viewpoint editing given an image of a single object. 
Similarly, works like DreamSparse~\cite{dreamsparse} also employ diffusion models to synthesize novel views on open-set categories.
Differently from our approach, these techniques are focused only on object orientation and rely on vast 3D shape datasets.
On the other hand, we investigate how to learn several continuous concepts (\eg illumination~\illumination, wing pose~\wing, dolly zoom~\dollyzoom) that can be directly used in text-to-image scenarios; \ie we don't generate an image to then change the orientation or illumination later, but instead we use \emph{Continuous Words} directly on text prompts.

\paragraph{Learning new concepts on diffusion models.}
With diffusion models being trained on unforeseen quantities in images and texts, a stream of work focused on adding specific concepts with very few data samples. 
For example, given a small set of images representing one particular object instance, textual inversion learns a new word embedding to describe the object, such that the word can be applied with new text prompts for image generation~\cite{textualinversion}. 
NETI~\cite{neti} extended the word embedding to a time-space conditioned neural mapper for better generation while preserving quality. 
Similarly, Dreambooth~\cite{ruiz2023dreambooth} aims to achieve the same goal, but by using a repurposed token rarely used in text and finetuning the entire diffusion model with an additional constraint to prevent generative loss. There are numerous subsequent works showing improvements on finetuning different layers/weights and by improving the training strategy \cite{customdiffusion,svdiff,cones}.
% Custom Diffusion~\cite{customdiffusion} shows that tuning just the $K$ and $V$ layers of cross attention would bring better quality than Dreambooth while allowing the model to learn
% multiple concepts at the same time. Other works like SVDiff \cite{svdiff} finetunes the singular values in the matrices.

\begin{figure*}[ht]
\centering
\includegraphics[width=\linewidth]{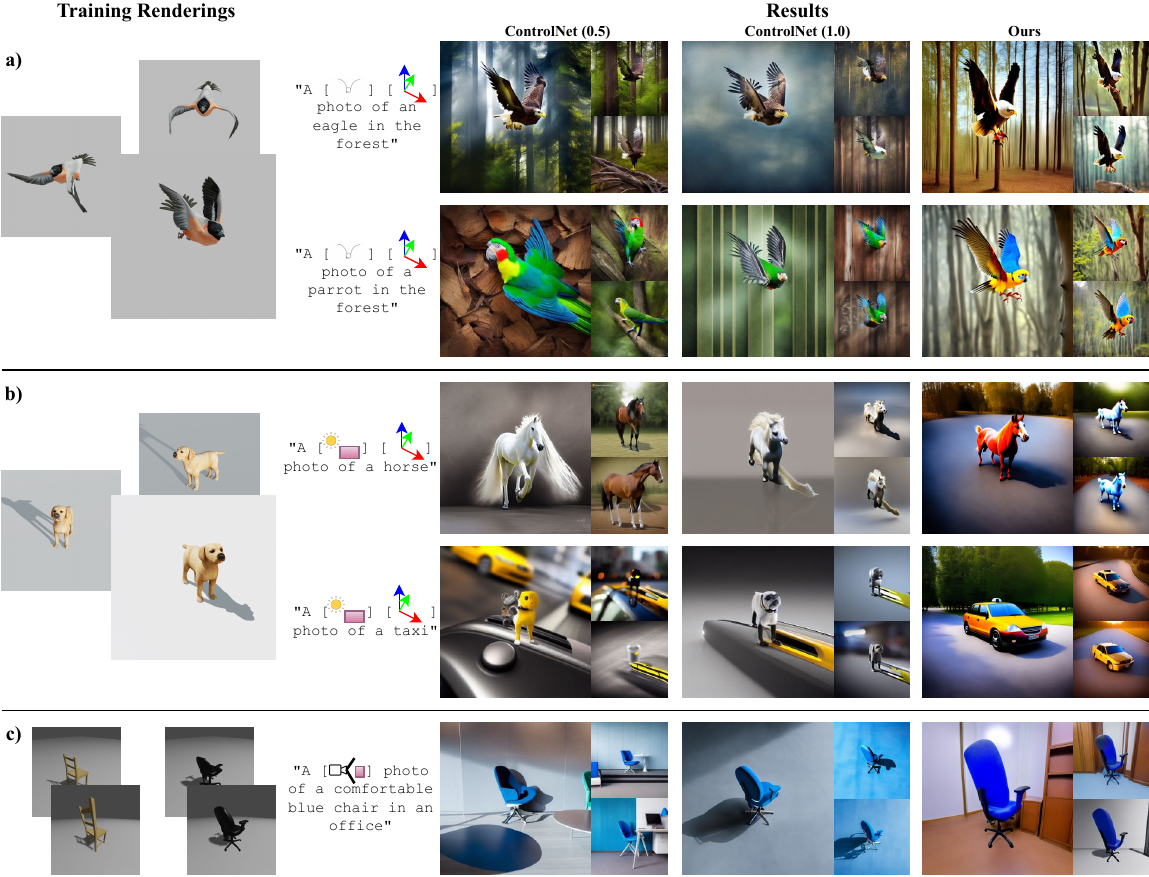}
% \vspace{-6mm}
\caption{
\label{fig:qualitative_comparison}
\textbf{Qualitative Comparisons.} We compare our Continuous 3D Words trained under three settings against ControlNet of various strengths. Note that the dollyzoom setup was trained with trained with multiple chair meshes, so we give additional ControlNet by manually picking the chair rendering that best follows the prompt (i.e., ``comfortable" and in ``the office").
}
% \vspace{-1mm}
\end{figure*}

Despite the advances in adding new personalized entities to existing models, few works focus on learning general concepts that can be applied to a variety of scenarios. 
A concurrent work, ViewNETI~\cite{viewneti}, is the first to learn viewpoints as a concept, but we hypothesise that the 3D awareness of large text-to-image diffusion models goes far beyond merely viewpoints, allowing us to associate and even create interactions with multiple 3D-aware concepts like illumination, pose and camera parameters \emph{at the same time}.
Our method, despite being trained only using a single mesh, shows superior generalization properties -- while trained to learn illumination and orientation from renderings of a single dog, we are capable of employing the learned concepts to generate cars, horses (Figure~\ref{fig:qualitative_comparison}), polar bears (Figure~\ref{fig:ablation}), lions (Figure~\ref{fig:teaser}) and so on.

\begin{figure}[t]
\centering
\includegraphics[width=\linewidth]{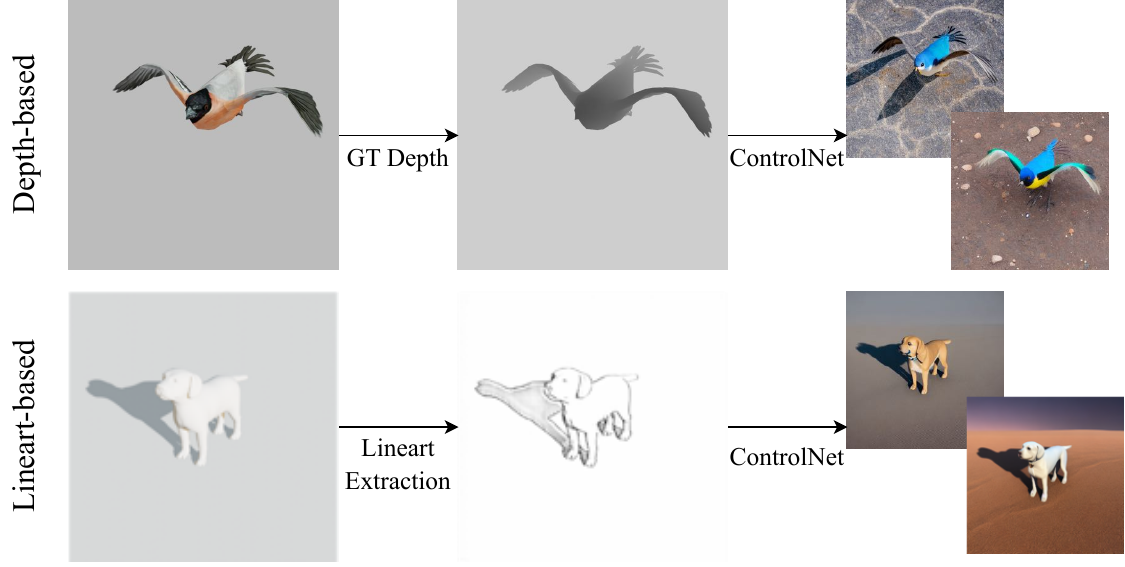}
% \vspace{-6mm}
\caption{
\label{fig:controlnet}
\textbf{ControlNet Augmentations.} Depth ControlNet is used for attributes creating direct shape changes. Lineart ControlNet is applied for more subtle changes that cannot be reflected by depths (e.g., illumination).
}
% \vspace{-1mm}
\end{figure}

\begin{figure}[ht]
\centering
\includegraphics[width=0.85\linewidth]{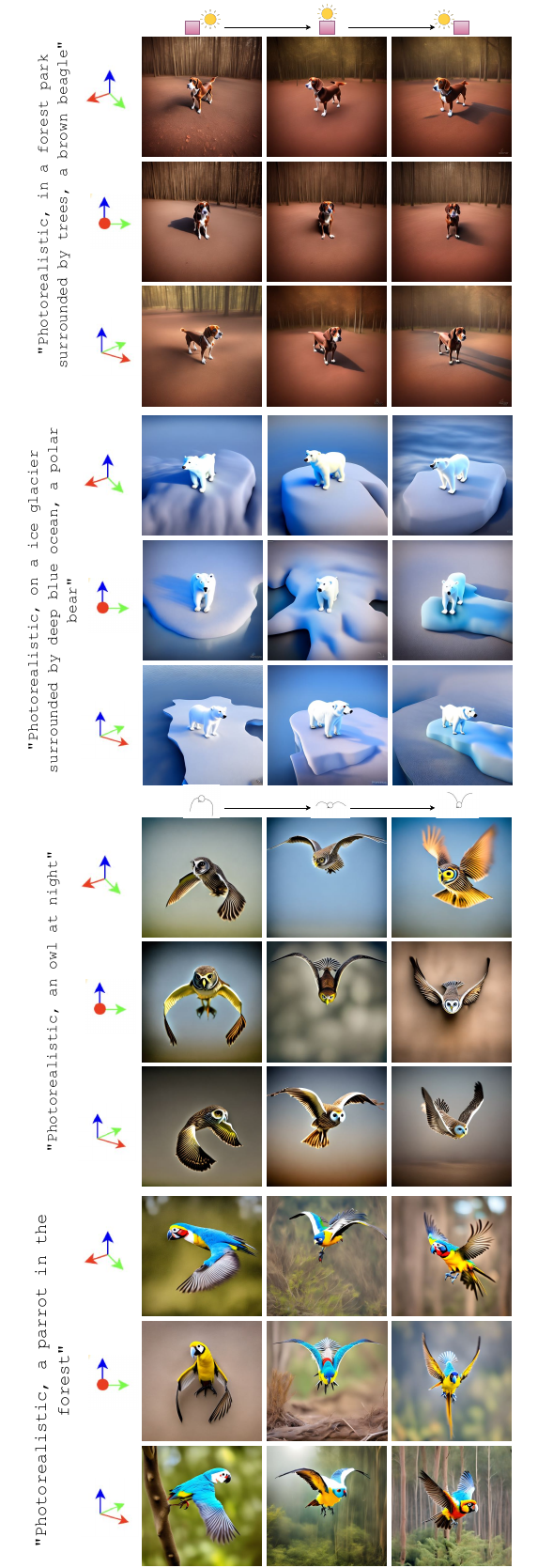}
% \vspace{-6mm}
\caption{
\label{fig:multi-concept_control}
\textbf{Disentangling Multiple Attributes.} We show four examples of controlling multiple Continuous 3D words in addition to text descriptions. The first 6 rows were trained with a single golden retriever mesh, while the bottom 6 were trained with a single animated dove.
}
\vspace{-8mm}
\end{figure}

\section{Method}
\label{method}

\subsection{Preliminaries}
We define an image $I$ that captures object $O$ from category $C$ as a function of several attributes $I = f(a_{1}, a_{2}, a_{3},..., a_{n})$, where $a_{i}$ belongs to a vast set of image attributes $\mathcal{A}$: shape, material reflectivity, rotation/translation, camera intrinsic/extrinsics, shape deformations, etc. 
Some of these components can be translated to other categories while others cannot, so for simplicity we assume they only work for a single category.
In the experimental section of this study, we will demonstrate that the definition of category for
some attributes is rather loose and the user is capable of generating images with \emph{continuous words} depicting objects very different from the ones seen during training.
Notice that images annotated with the attributes we are interested are very rare, so the models
do not have a very precise knowledge of them, except for what is already described
in text-image pairs. 

Given an image set $I^O$ with images capturing an object $O$, previous methods like Dreambooth~\cite{ruiz2023dreambooth}, Custom Diffusion~\cite{customdiffusion}, or Textual-Inversion~\cite{textualinversion} aim to minimize the following objective:
\begin{equation}
 \mathbb{E}_{\hat{I}_{\epsilon,i}, T_O, I_i \in I^O}
 \bigg[
     \norm{S_{\theta}(\hat{I}_{\epsilon,i}, P(T_O)) - I_i}^2_2
 \bigg],
\end{equation}
where $S_{\theta}$ is a Text-to-Image diffusion model~\cite{stablediffusion}, $\hat{I}_{\epsilon,i}$ is a noised image $\alpha_t I_{i} + \sigma_t \epsilon$ with noise $\epsilon$ and noise schedulers $\alpha_t, \sigma_t$. $P(T_O)$ is the prompt condition which contains a token embedding $T_O$ used as an identifier of object $O$. In practise $P(\cdot)$ is the text encoder from CLIP.
% \todo{We should add a short sentence explaining how P is implemented in practice -- a function of 
% the text encoder etc}.
The fine-tuned network $S_{\theta}$ can then generate new images containing $O$ when given a new prompt condition $P'(T_O)$ and some Gaussian noise. 
Unlike previous methods, our goal is not to add concepts representing specific objects, 
but rather have them describe some attributes $a_{i} \in \mathcal{A}$, by learning to disentangle them using as few objects as possible within $C$ -- most of the time just one object suffices. 
Our model can also be easily extended to allow control of multiple attributes \emph{at the same time}. 
% I don't think we explore this in the paper, so will leave this out.
%Consequently, given one single image of a new object $I^n$, $n \in C$, we can generate new images of $n$ while providing fine-grained control over $a_{i}$.

\subsection{Continuous Control}
A naive way to control an attribute $a$ is to use some realistic rendering engine to generate images of the available objects that have the same value $a = x$, and then apply similar approaches to previous works by assigning a token $T_x$ to identify images with this particular value. 
This is not ideal, however, since $a$ is often continuous and have infinitely many values -- we would require an unfeasible number of tokens to gain fine-grained control over these attributes.

Therefore, we propose to instead learn a continuous function $g_{\phi}(\mathbf{a}) : \mathcal{D} \rightarrow \mathcal{T}$ that maps a set of attributes from some continuous domain $\mathcal{D}$ to the token embedding domain $\mathcal{T}$.
We use positional encoding to first cast each attribute $a \in \mathbf{a}$ to a higher frequency space before feeding into the function, which is represented by a very simple 2-layer MLP.
The output of this network is named \emph{Continuous 3D Word} and will allow
users to easily control continuous attributes from text prompts augmented by these tokens.
Finally, our training objective can then be formulated as:
% \red{Shouldnt it be $g_{\phi}^{a_{cl}} x \mapsto \mathcal{T}$?} \red{you might want to introduce the notation $\hat{I}_\epsilon = \alpha_t I_{a_{cl}} + \sigma_t \epsilon$ for a noisy image. It would make it fit to a line, plus how an imge is noised is not our focus here.}
\begin{equation}
 \arg \min_{\theta, \phi} \mathbb{E}_{\hat{I}_{\epsilon, \mathbf{a}}, \mathbf{a}}
\bigg[
    \norm{S_{\theta}\big(\hat{I}_{\epsilon, \mathbf{a}}, P(g_\phi(\mathbf{a}))\big) - I_{\mathbf{a}}}^2_2 
\bigg].
 \label{eq: continuous_control}
\end{equation}

\subsection{Disentangling Object Identity and Attributes}
In practice, when we only utilize a single object to learn attributes, a degenerate solution occurs when directly optimizing (\ref{eq: continuous_control}), where $S_\theta$ treats the same object with different values for $a$ as different objects.
% \red{Is this beause the naive strategy involves training with a single object? We have not say so at this point}.
This hinders the generalization capability of changing the attribute when given an image of a new object.
To this end, we propose two strategies, one during training and one during inference, to disentangle the object identity and individual attributes.
\vspace{4pt}

\noindent\textbf{Training: Learning identity and attributes.}
We provide a simple regularization by explicitly forcing the model to use the same identifier for images representing the same object. The objective can thus be formulated as:
\begin{equation}
 \arg \min_{\theta, \phi} \mathbb{E}_{\hat{I}_{\epsilon, \mathbf{a}}, \mathbf{a}}
\bigg[
    \norm{S_{\theta}\big(\hat{I}_{\epsilon, \mathbf{a}}, P(T_O, g_\phi(\mathbf{a}))\big) - I_{\mathbf{a}}}^2_2 
\bigg].
 \label{eq: disentangle}
\end{equation}
Optimizing both $T_O$ and $g_\phi(\mathbf{a})$ concurrently proved to be difficult from our experiments (see Section \ref{sec: ablation}). We propose a two-stage training strategy as depicted in Figure \ref{fig:pipeline}. 
For every available image in $I^O$ with varying $\mathbf{a}$, we first use the same prompt condition $P(T_O)$ to associate them to the same object, then learn the diffusion model parameters $\theta$ and our Continuous 3D Words MLP $g_\phi$ by using the prompt condition $P(T_O, g_\phi(\mathbf{a}))$. 
% \red{I know they have been defined separately, but it would help the reader to recap in one sentence what are the variables we optimize : $\theta$ for the LoRA and $\phi$ for the continuous words.}
\vspace{4pt}

\noindent\textbf{Inference: Negatively prompting object identifier.}
To further the disentanglement of attributes against object identities. We propose a simple trick during inference by adding the object identity as a negative prompt. 
Specifically, for each sampling step, we swap the null-text embedding used by classifier-free guidance with $T_O$.
Intuitively, since our training happened mostly using $O$, we want to disincentivize the model to generate images containing such object.

\subsection{ControlNet Augmentation}
\label{sec:controlnetaug}
To prevent the fine-tuning process from overfitting to a simple white backgrounds and pre-defined object textures, we augment the backgrounds and textures in the rendering process.
However, directly doing this in simulation engines is time consuming specially if one is targetting realistic scenes. 
Thus, we propose an automated solution by utilizing pretrained ControlNets.

Figure \ref{fig:controlnet} shows two types of ControlNet augmentations we used in our framework. For attributes that can be directly reflected on shape changes (e.g., wing pose), we directly render the ground-truth depth maps to use as the condition for ControlNet. On the other hand, for attributes that cannot be reflected directly from depths (e.g., illumination), we first render the images without textures, then use a lineart extractor to obtain a ``sketch" of the image. This captures subtle changes such as shades and shadows in the pixel space, which can then be used as the condition for Lineart ControlNet.
% \todo{This is a very good explanation. It should be repeated in the caption of Figure 3 (different, more succinct words, of course)}.

We add additional prompts describing the object appearance and background during the ControlNet generation.
It is important to note that prompts deviating away too much from the original mesh category would lead to degenerate images, so our prompts are in general very simple and straightforward (details in supplemental material).
We include the ControlNet generated images as a small set of data augmentation, and we use the same prompt which we use to guide ControlNet generation to guide our Stage 2 fine-tuning.

\section{Experiments}
We use off-the-shelf Stable Diffusion v2.1~\cite{stablediffusion} as the backbone of our method.
We resort to the recent Low-Rank Adaptation (LoRA)~\cite{lora} for the fine-tuning of the denoising U-Net and text encoder, allowing us to train on a single A10 GPU occupying roughly 16GB of memory. 
Thanks to the low-rank optimization, our models have a very small size (approximately 6MB).
Training time varies by the complexity of the single/multiple attributes to learn, but falls within 15k to 20k steps, which generally takes around 3-4 hours in a single GPU.
For ControlNet augmentation, we use the official implementation of ControlNet v1.1~\cite{controlnet}. 

We implement our Continuous 3D words under five different attribute settings. 
For single attributes we implement 1) illumination~\illumination using a single dog mesh, 2) wing pose~\wing using a single animated dove mesh, and 3) Dolly zoom~\dollyzoom with five Pix3D chairs~\cite{pix3d}. 
For multi-concepts, we train 4) illumination and object orientation \illumination+\orientation using a single dog mesh and 5) wing pose and orientation \wing+\orientation using a single animated dove mesh. 
Settings 1) and 4) use lineart images~\cite{lineart} while the others use depth map to compute the ControlNet background augmentation (see Section~\ref{sec:controlnetaug}).

\subsection{Comparison with Baselines}

\noindent\textbf{Baseline Design.}
We design a very competive baseline that enables fine-grained attribute control in image generation
by combining the mesh training data we used in our experiments, a rendering engine and ControlNet~\cite{controlnet}.
Specifically, we take a \textit{novel} text prompt for a \textit{training} object, and grab a corresponding condition map in the training set rendered with the intended attributes.
For example, if the prompt is \texttt{a} \wing \texttt{eagle flying in a forest}, we select
the frame of the dove mesh that corresponds to the user-prescribed wing pose~\wing, render
its depth map using a rendering engine and pass it through ControlNet with the same prompt
but removing the \emph{continuous 3D word}.
The strength of the ControlNet guidance is a critical hyperparameter --- increase in strength could increase the accuracy of reflecting the attribute but decrease the robustness to generalize to the text-prompt intended object.
Therefore, we present the ControlNet baseline with both full and half strength in terms of guidance.
We also explored an interpolation of null-text embedding~\cite{mokady2022null} baseline,
but the results failed to capture even the simplest attributes, so they were omitted from our analysis.
\vspace{4pt}

% When controlling only the pose attribute, where we can define the scenario and absolute pose of an object in an image, we also compare with Zero-1-to-3 [CITE Zero123]. Note that our continuous vocabulary does not restrict the appearance of the object as long as it follows the viewpoint direction, unleashing greater generalization capabilities, it is inherently different to the problem setting of Zero-1-to-3 rotating a pre-existing image. Therefore, we slightly amend Zero-1-to-3 by first using our method to obtain the object at canonical pose, then rotate it with Zero-1-to-3.

\begin{table}[t!]
    \centering
        \resizebox{\linewidth}{!}{
        
    \begin{tabular}{lcccc|c}
    \toprule
    &\multicolumn{5}{c}{\textbf{User Preference (\%) $\uparrow$}} \\
    \cmidrule(lr){2-6}
          \textbf{Method} & \wing & \illumination & \wing/\orientation & \illumination/\orientation & \textbf{Avg.}\\
     \cmidrule(lr){2-6}

            ControlNet (1.0)    & \cellcolor{myyellow}28.3\% &  16.2\%  & \cellcolor{myyellow}35.0\%  &15.0\%& \cellcolor{myyellow}23.6\%\\
            ControlNet (0.5)    & 10.0\% &  \cellcolor{myyellow}28.8\%  & 12.5\%  &\cellcolor{myyellow}32.5\% & 21.0\%\\
                        Ours    & \cellcolor{myred}61.7\% &  \cellcolor{myred}55.0\%  & \cellcolor{myred}52.5\% & \cellcolor{myred}52.5\% &\cellcolor{myred}55.4\%\\
    \midrule
    &\multicolumn{5}{c}{\textbf{Average User Ranking $\uparrow$}} \\
    \cmidrule(lr){2-6}
      \textbf{Method} & \wing & \illumination & \wing/\orientation & \illumination/\orientation & \textbf{Avg.}\\
     \cmidrule(lr){2-6}

            ControlNet (1.0)    & \cellcolor{myyellow}$2.07\pm 0.70$    & $1.49\pm 0.76$ & \cellcolor{myyellow}$2.20\pm 0.68$& $1.60\pm 0.74$ &\cellcolor{myyellow} $1.84 \pm 0.72$\\
            ControlNet (0.5)    & $1.38\pm 0.66$    & \cellcolor{myyellow}$2.11\pm 0.67$ & $1.38\pm 0.70$& \cellcolor{myyellow}$2.06\pm 0.76$& $1.73 \pm 0.70$\\
                        Ours    & \cellcolor{myred}$2.55\pm 0.62$    & \cellcolor{myred}$2.40\pm 0.73$ & \cellcolor{myred}$2.43\pm 0.67$& \cellcolor{myred}$2.33\pm 0.77$ & \cellcolor{myred}$2.43 \pm 0.70$\\
    \bottomrule
    \end{tabular}
    }
    % \vspace{-2mm}
    \caption{\textbf{User study results.}
    We asked users to rank three images according to their preference and how well they followed the given conditions -- text prompt
    and one or two continuous controls.
    The controls were described by representative images (\ie arrows for orientation, shaded sphere for illumination, \etc).
    \emph{See supplemental material for more details}.
    We evaluate four different control types: wing pose~\wing, illumination~\illumination, wing pose~\wing + orientation~\orientation, and illumination~\illumination + orientation~\orientation.
    Cells colored as {\color{myred}{red}} and {\color{myyellow}{yellow}} represent the best and second best method, respectively.
    Our method was selected as the favorite for the majority of users in all evaluated setups.
    }
    \vspace{-4mm}
    \label{tab:user Preference}
    \end{table}
    
\noindent\textbf{Quantitative Results.} Due to the complexity and abstract nature of the attributes we are analyzing,
automatically measuring whether a generated image reflects a set of values is a considerable challenge.
Thus, following previous papers~\cite{ruiz2023dreambooth,prompt2prompt,controlnet}, we create a user study to evaluate the quality of the generated images while controlling several attributes.
For each setting 1, 2, 4, and 5, we randomly sample a set of 3D conditions and prompts, generating over 60 questions per setting.
Similarly to the user studies above, we then invite 20 participants, showing them all the given constraints we want the image to pertain (prompts and attributes), and ask them to rank each image from best to worst.

Table \ref{tab:user Preference} shows the results in both the percentage of user preference (image chose as the best one) and the overall ranking. Best results are highlighted as {\color{myred}{red}}. Our \emph{Continuous 3D Words} won the majority of votes (over 50\%) in all analyzed scenarios.
Interestingly, the second best (as highlighted in {\color{myyellow}{yellow}}) alternates between the two guidance strengths of ControlNet. For wing-pose based controls, having a weaker control diminishes the strong prior offered by the depth map, resulting in inaccurate poses. Conversely, strong guidance for illumination forces ControlNet to generate objects around the shadow outline (further shown in Figure \ref{fig:qualitative_comparison}).
Differently from the baselines, our training strategy is one-size-fits-all without any hyperparameter required.
\vspace{4pt}

% \noindent\textbf{Extending to objects significantly different}
% The benefits of our method becomes even more apparent when applying our learnt attributes to objects with a significantly different semantic to the training data. \tim{Depends on if we have space, we can put an additional table only highlighting ones deviating for illumination.}

\noindent\textbf{Qualitative Analyses}
We present a detailed comparison of Continious 3D Words against ControlNet of different guidance strengths in Figure~\ref{fig:qualitative_comparison}. We show the results under three training settings : a) wing pose and orientation, b) illumination and orientation, and c) dollyzoom. We observe that the dollyzoom setup was a harder concept to train and had to be done using five chairs. 
We also manually ``helped'' the ControlNet baseline by manually picking the chair that best followed our text prompt as the condition image.
More importantly, the ControlNet baselines significantly deteriorate when the prompt contains 
elements that were not present in the training data.
For example, even when trained on a single dog mesh, our method can learn illumination and orientation
attributes that can be used to generate horses and taxis (see row \emph{b)}, Figure~\ref{fig:qualitative_comparison}).

%From Figure~\ref{fig:qualitative_comparison}, we can see that both guidance strengths have their drawbacks.
%Having strength of 1.0 can usually reflect the attributes given that the text prompt doesn't deviate drastically from the training data. However, the quality of the image is very often lost as the shape has to strictly follow the shape of the conditioning image. The result worsens when the difference between prompt and training mesh enlarges (e.g., ControlNet 1.0 generates a dog instead of a taxi). On the other hand, while decreasing the guidance strength to 0.5 often increases the generation quality, the images fail to properly reflect the imposed conditions (e.g., the horse from ControlNet 0.5 loses the illumination).

\subsection{Multi-Concept Control}

Just like sentences where we can encompass multiple words, but each disentangled from one another when controlling the image generation, our Continuous 3D Words can do the same. We show four examples in Figure \ref{fig:multi-concept_control}, two from \illumination and \orientation, and two \wing and \orientation, where we can keep one attribute fixed but change the other without sabotaging the quality of image generations. They can be jointly used with complex prompts describing the background and object texture. Moreover, while all these words are learned from a single mesh, we can easily transfer the attribute to objects with fairly close semantics (e.g., a labrador mesh to a polar bear, or a dove mesh to a parrot).

\begin{figure*}[t]
\centering
\includegraphics[width=\linewidth]{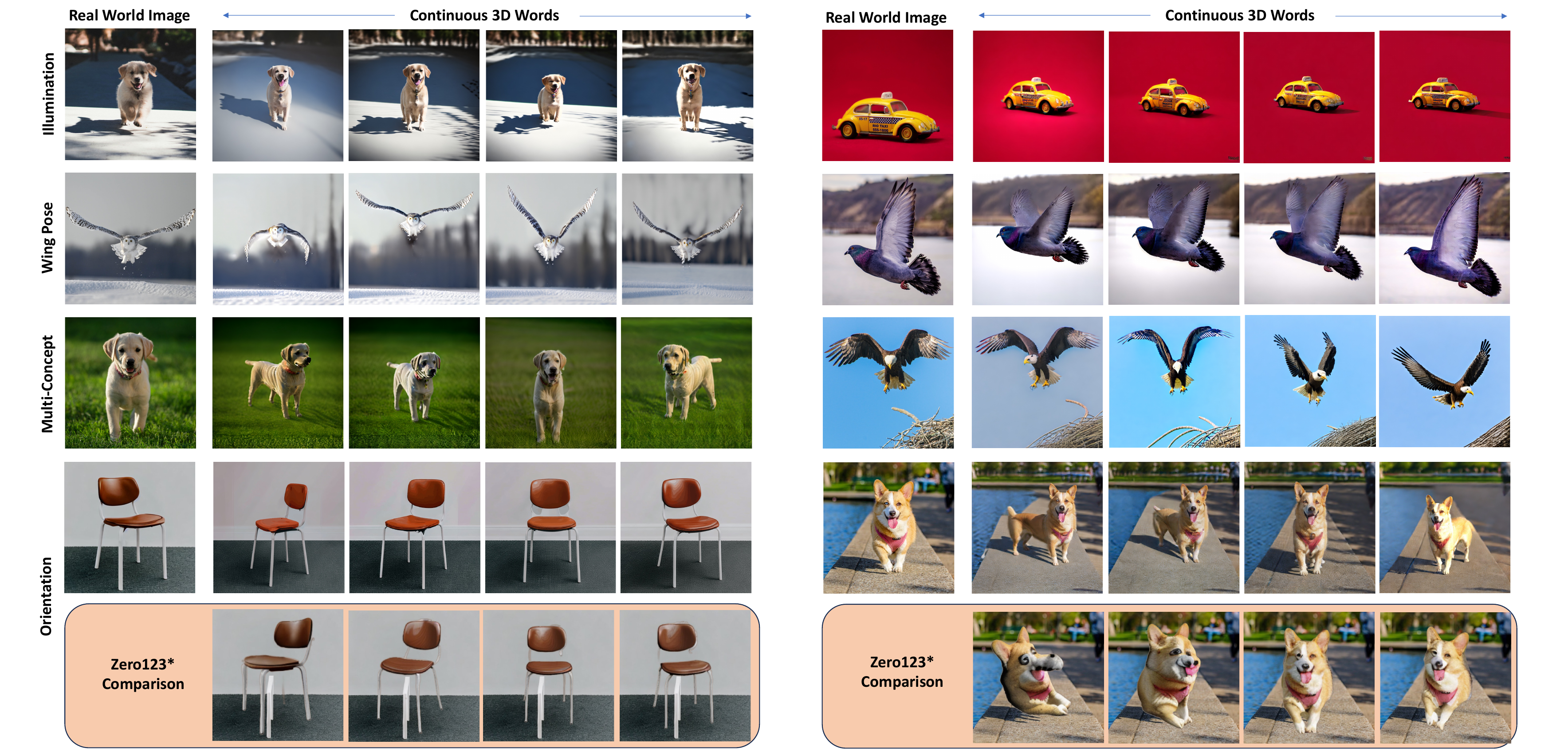}
% \vspace{-6mm}
% \vspace{-3mm}
\caption{
\label{fig:realworld}
\textbf{Real World results \& Comparison w/ Zero1-to-3.} 
We learn tokens representing single images and use along with
\emph{Continuous 3D Words} for image editing.
The learned image token encodes most of the image appearance, while the \emph{Continuous 3D Word}
modifies its relevant aspects.
Zero123, on the other hand, tends to yield images where the modified object is deformed or not properly harmonized.
}
% \vspace{-3mm}
\end{figure*}
\subsection{Real World Image Editing}

Our Continuous 3D Words can be directly applied to real world images to perform editing.
To do so, we simply have to encode a real world image to a rare token via Dreambooth \cite{ruiz2023dreambooth}.
Then, we only have to use that token in conjunction with our Continuous 3D Words to generate the edited image. 
We show 8 examples in Figure \ref{fig:realworld}, changing \illumination, \wing, \orientation.
As we can see, the Dreambooth token preserves most of the image appearance, while our Continuous Words understands and brings edits to the main subject.

\vspace{4pt}\noindent\textbf{Comparing with Zero-1-to-3} While our approach is is focused on enhancing text-to-image, given the capabilities of real-world image editing, we can use the same setup to provide a comparison with \emph{only} with object orientation changes (Results in Figure \ref{fig:realworld}).
Notice that Zero1-to-3 \cite{liu2023zero1to3} operates in foreground-only images so we also had
to segment the object \cite{rembg}, inpaint the background \cite{stablediffusion} and place the novel orientation back to into the image. 
Each one of these steps contain errors that, when compounded, hurt
the quality of the final result.
More importantly, our method allows controls beyond orientation changes
without relying on massive 3D datasets.

\section{Discussion and Limitations}

\begin{figure}[t]
\centering
\includegraphics[width=\linewidth]{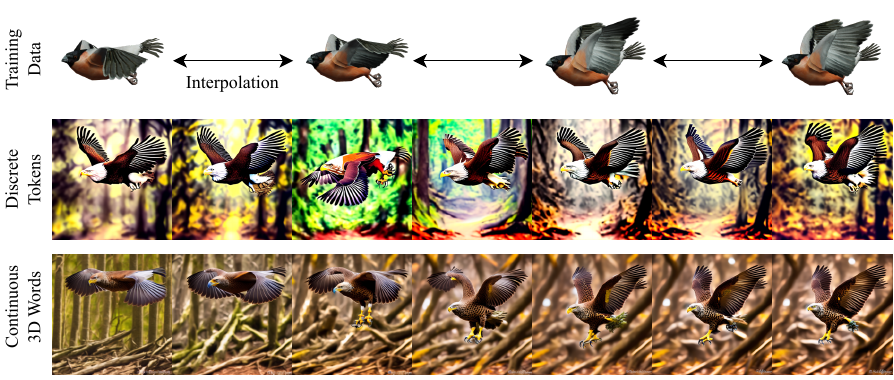}
% \vspace{-6mm}
\caption{
\label{fig:interpolation}
\textbf{Interpolating continuous 3D words.} We present two results of interpolation, one with discrete tokens and one with our Continuous 3D Words. Our method preserves the attributes better and enables interpolation between two values.
}
% \vspace{-1mm}
\end{figure}

\noindent\textbf{Why Not Discrete Tokens?} 
The benefits of fitting a single MLP instead of multiple tokens on an attribute with different values are two-fold. First, the MLP learns a continuous function, allowing us to interpolate between two training data samples whereas fitting multiple tokens leads to two datapoints as two discrete mappings.
Second, finetuning a model to learn multiple custom tokens simultaneously is very hard.
These benefits are illustrated in Figure \ref{fig:interpolation}.
We show a comparison of learning a single \emph{Continuous 3D Word} \wing and 18 discrete tokens for different values of a wing pose.
Both were learned with the same training method (2-stage with ControlNet augmentation).
During inference, we present the generated results of three attribute values that are present in the training set, as well as the results when you interpolate the value in between. Interpolation is straightforward for our case where we just input the intermediate value into our Continuous 3D Words MLP.
For discrete token, we take the interpolation of the two nearest discrete bins.
Notice that the discrete tokens have difficulty not only in interpolating results but also in learning all the concepts simultaneously. Notice how the wings of the eagle in the second row of Figure~\ref{fig:interpolation} do not follow the training data and sometimes generate a completely different pose (third column, second row).  
On the other hand, our images closely follow the pose prescribed by the user (top row) while
yielding appealing images even when the values were not seen during training (columns 2, 4 and 6; second row).
\vspace{4pt}

\begin{figure}[t]
\centering
\includegraphics[width=\linewidth]{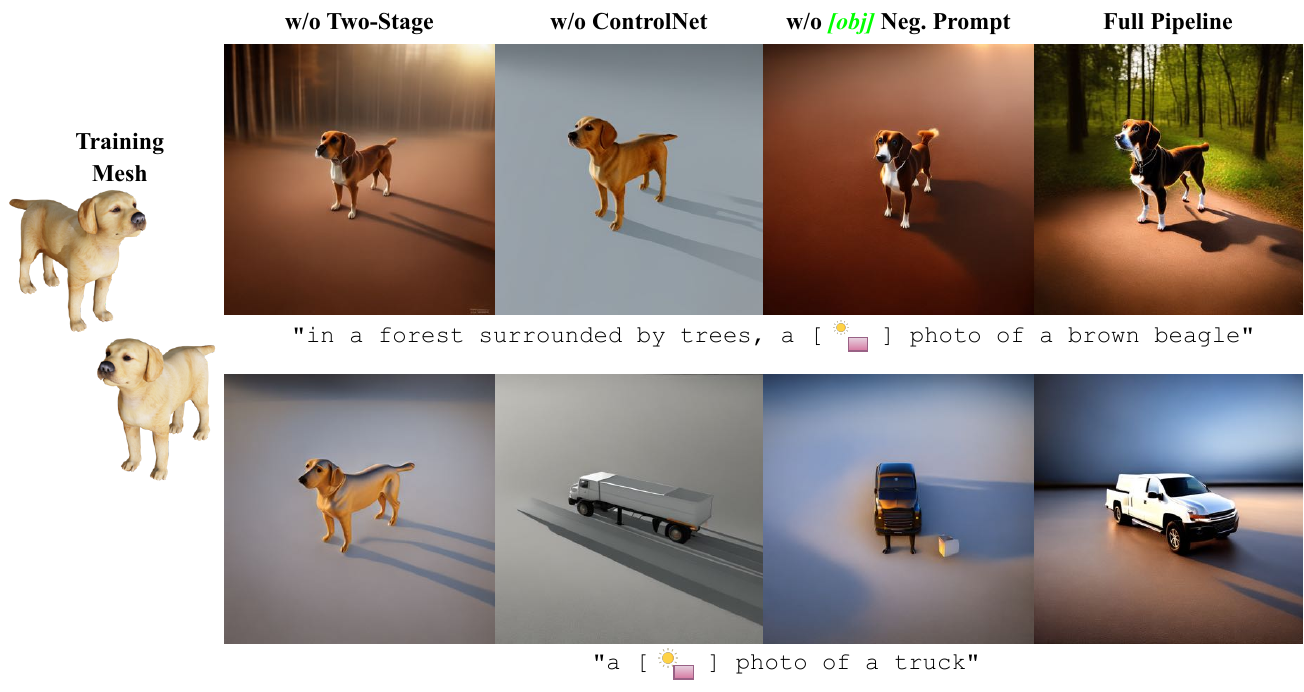}
% \vspace{-6mm}
\caption{
\label{fig:ablation}
\textbf{Ablation Study.} We present our ablations for: w/o two-stage training, w/o ControlNet augmentation, w/o \textit{\textcolor{green}{[obj]}} as negative prompt, and our full pipeline. Notice that, when the prompt deviates a lot from the training data (truck in prompt, mesh of a dog for training), the ablated version fails to follow the prompt.
Without two-stage training, it ignores the prompt and creates a dog; without the other
parts it yields deformed shadows and dog-shaped trucks.
}
% \vspace{-1mm}
\end{figure}

\noindent\textbf{Ablation study.} \label{sec: ablation}
Figure \ref{fig:ablation} shows an ablation of our training strategy.
We remove each component of our training strategy (w/o two-stage training, w/o ControlNet augmentation, w/o \textit{\textcolor{green}{[obj]}} as negative prompt) and compare it with our full pipeline. Two examples are presented: one where the prompt is similar to the training and another
one where it is significantly different. 
Without the two-stage training, the model fails to disentangle object identity with our attributes, hindering the generalization capability to new objects.
This is particularly noticeable for the bottom row when the text prompt is a \texttt{truck} but a dog similar to the training mesh is generated when the two-stage training is removed.
Without ControlNet, the finetuning process often overfits to the background training renderings, resulting in an inability to generate realistic backgrounds.
Finally, adding \textit{\textcolor{green}{[obj]}} as the negative prompt serves as a minor improvement in further disentangling both the backgrounds and object shape seen during training, resulting in a more aesthetic image.
\vspace{4pt}

\begin{figure}[t!]
\vspace{-4mm}
\centering
\includegraphics[width=\linewidth]{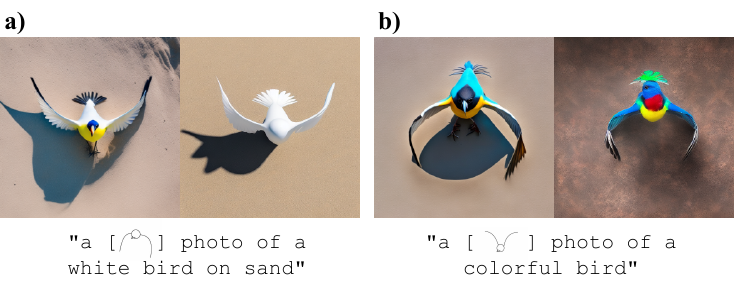}
% \vspace{-6mm}
\caption{
\label{fig:user_study}
\textbf{Condition v.s. Accuracy in User Study.} a) shows two images generated by the prompt ``a \wing photo of a white bird on sand". b) shows two images generated by the prompt ``a \wing photo of a colorful bird". Left is ours, right is ControlNet (1.0). Users  preferred right over left for both.
}
% \vspace{-1mm}
\end{figure}

\begin{figure}[t!]
\vspace{-4mm}
\centering
\includegraphics[width=\linewidth]{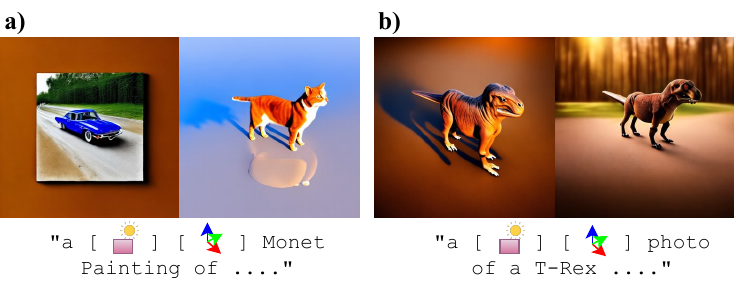}
\vspace{-6mm}
\caption{
\label{fig:failure}
\textbf{Failure cases.} a) shows two images generated by the prompt ``a \illumination \orientation Monet Painting of .....". b) shows two images generated by the prompt ``a \illumination \orientation photo of a T-Rex". Both yielded suboptimal results.
}
\vspace{-4mm}
\end{figure}
\noindent\textbf{Condition v.s. Generated Accuracy in User Study.} During our user study, we realize that occasionally users select the image which more strictly follows the exact conditions over the image that is more physically probable. For example, in both cases a) and b) shown by Figure \ref{fig:user_study}, the head of the bird is generated poorly, but users preferred them as they are ``whiter" and ``more colorful". 
\vspace{8pt}

\noindent\textbf{Failure Cases.}
We present in Figure \ref{fig:failure} two examples of typical failure cases in our results. First, our model currently fails on more difficult scenarios where the style is given by the text prompt. In a), the image cannot fully reflect the ``Monet painting" style imposed by our prompt. Second, the generated object may sometimes still overfit to the training set. In b), the T-Rex had four feet on the ground instead of standing with two claws in air -- an attribute that is similar to the training dog mesh used to learn illumination.  

% the closest proximity 

% Due to the open challenge of quantita
% \begin{figure*}[!t]
%     \centering
%     \includegraphics[width=\linewidth]{author-kit-CVPR2024-v2/figures/fig4_gen_comparison.pdf}
%     \vspace{-10pt}
%     \caption{{\textbf{Complex composition fine-tuning results.} Given images of a new concept (target images shown on the left), our method and composed data can generate images with the concept in complex contexts as well as boosting the existing methods (e.g., Custom Diffusion and Dreambooth) to improve. (\daniel{change ours to our method + our data, and ...})
%     } }
%     \label{fig:results_complex}
%     \vspace{-5pt}
% \end{figure*}

\section{Conclusion}
We presented \emph{Continuous 3D Words}, a framework that allows us to learn 3D-aware attributes 
reflected in renderings of meshes as special words, which can then be injected into the text prompts 
for fine-grained text-to-image generation.
We made an extensive study on learning both single and multiple continuous words and show that we can control challenging attributes.
With the lightweight design and promising results, we hope that this work opens up interesting applications in the vision community to create their own 3D words with a single mesh and an accessible rendering engine.
\vspace{4pt}

\noindent\textbf{Future work.}
Identifying which data needs to be used for specific attributes and training multiple
models for each of them is a cumbersome task.
On the other hand, as the amount of 3D data available significantly increases, we believe that
an interesting direction is to train general models that handle multiple attributes on their own,
without the need to train attribute-specific networks.

{
    \small
    \bibliographystyle{ieeenat_fullname}
    \bibliography{main}
}
\clearpage
\maketitlesupplementary

\section{ControlNet Augmentation Prompts}
\label{sec:controlnet_supp}

Both our depth and lineart ControlNet augmentations for training had to be carefully designed as too large of a deviation may lead to wrongly synthesized images. We describe the prompts used for the ControlNet augmentations for our settings 1) wing pose, and 2) dollyzoom, and 3) illumination.

\noindent\textbf{Wing pose \wing.} Since the correct position of the wings is particularly important for this case, we engineered the prompts that helped the most in reflecting both wings during generation, which are \texttt{\{with two wings, flying\}}. We also added two time-of-day prompts to increase the variety of backgrounds. Therefore, the overall prompts are generated randomly with: \texttt{a bird \{with two wings, flying\} on a \{rainy, sunny\} day}.

\noindent\textbf{Dollyzoom \dollyzoom.} As there already comprises 5 types of chair inside the training dataset for dollyzoom we focus on augmenting the background with ControlNet. Our overall prompts are generated by: \texttt{a chair \{in the Acropolis, in a forest, under the snow, on a beach, in Times Square, in a department store\}}.

\noindent\textbf{Illumination \illumination.} We realize that Lineart ControlNet for shadow generation works best with ControlNet guidance 0.6. However, this weaker strength limits the ability for ControlNet to keep the shadow/illumination consistency if additional backgrounds are described (e.g., adding background descriptions like \texttt{in a forest} during Augmentation). Hence, our ControlNet augmentation only focuses on a variety of different dogs. Our overall prompts are generated by: \texttt{a \{white, gray, brown\} dog}.

\noindent\textbf{Multi-Concept Control.} When training multi-concept controls for by adding orientation to wing pose/illumination, we still use the same prompts as described above.

\section{User Study Examples}
\begin{figure}[t]
\centering
\includegraphics[width=0.8\linewidth]{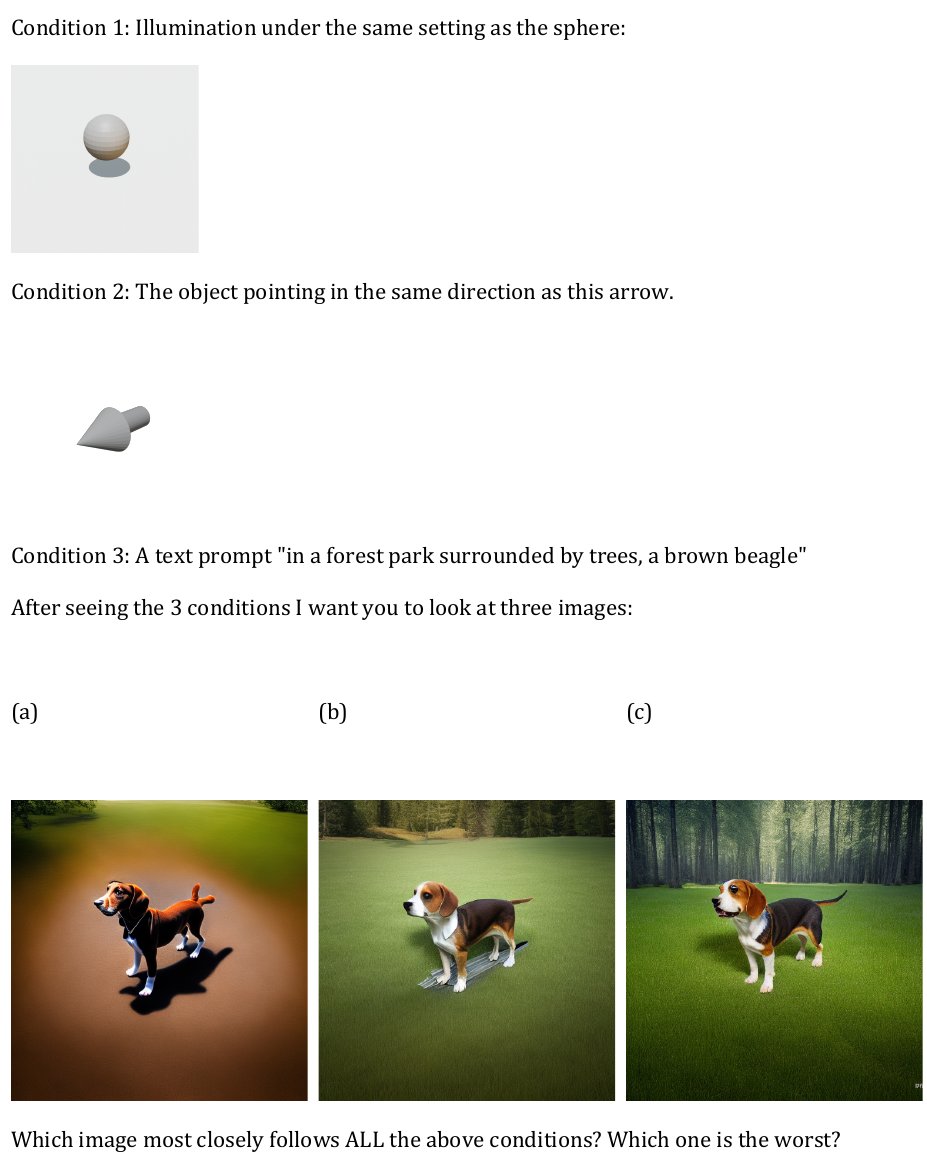}
% \vspace{-6mm}

\caption{ \label{pose_illum_surv}
Orientation/Illumination Survey.
}
\vspace{3mm}
\end{figure}

\begin{figure}[t]
\centering
\includegraphics[width=0.8\linewidth]{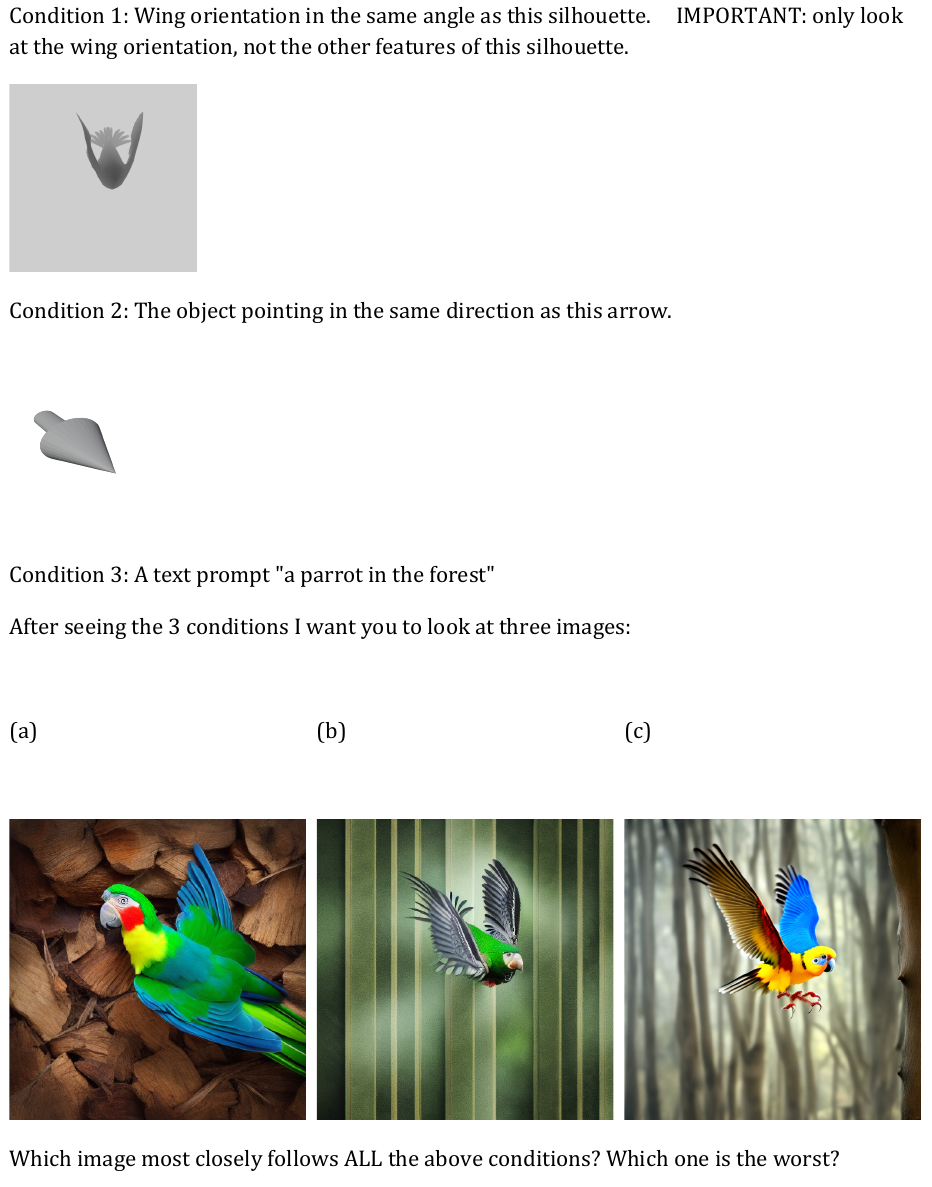}
% \vspace{-6mm}

\caption{ \label{pose_wing_surv}
Orientation/Wing Pose Survey.
}
% \vspace{-1mm}
\end{figure}

We provide two examples, from object \wing and \orientation and \illumination and \orientation, respectively as shown in Figure \ref{pose_illum_surv} and \ref{pose_wing_surv}.

\subsection{Prompt Selection.} Our prompts used for user study comparison are designed such that objects from very close to very far proximity from the training mesh are tested. 
For illumination \illumination , our comparisons involve \texttt{beagles} (high proximity to the dog mesh),\texttt{horse} (medium proximity to the dog mesh), and \texttt{taxi, rockets} (low proximity to the dog mesh).

Similarly, for wing pose \wing , our comparisons involve \texttt{bird with black head} (similar color to the training bird mesh) to \texttt{eagle/parrot in the forest} (different identity with additional background descriptions). 

\begin{figure}[t]
\centering
\includegraphics[width=1.02\linewidth]{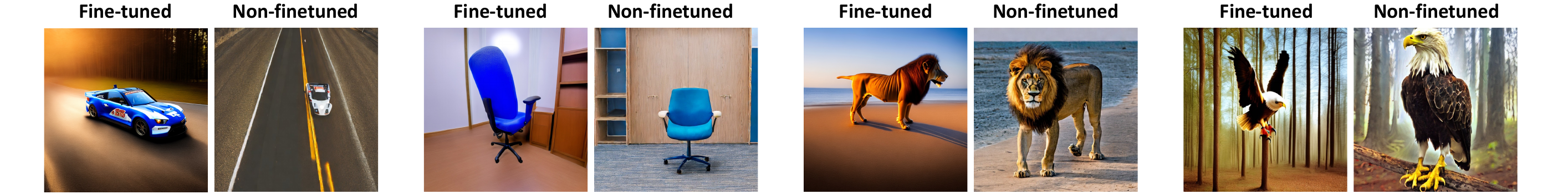}
% \vspace{-7mm}
\caption{
\label{fig:texture}
\textbf{Non-finetuned Stable Diffusion Comparison.} We compare the generated images using the same prompts with/without finetuning the diffusion model to ensure there is no significant texture difference. 
}
% \vspace{-5mm}
\end{figure}

\noindent\textbf{Comparison Against Non-finetuned Stable Diffusion.} We provide 4 example comparisons using the same prompts for text-to-image generation against unfinetuned stable diffusion (Figure \ref{fig:texture}) to evaluate the texture differences before/after fine-tuning. While there may be a slight texture difference (potentially due to the limited single-mesh training set), we believe our newly generated image are still of high quality.
% 
% Having the supplementary compiled together with the main paper means that:
% 
% \begin{itemize}
% \item The supplementary can back-reference sections of the main paper, for example, we can refer to \cref{sec:intro};
% \item The main paper can forward reference sub-sections within the supplementary explicitly (e.g. referring to a particular experiment); 
% \item When submitted to arXiv, the supplementary will already included at the end of the paper.
% \end{itemize}
% % 
% To split the supplementary pages from the main paper, you can use \href{https://support.apple.com/en-ca/guide/preview/prvw11793/mac#:~:text=Delete%20a%20page%20from%20a,or%20choose%20Edit%20%3E%20Delete).}{Preview (on macOS)}, \href{https://www.adobe.com/acrobat/how-to/delete-pages-from-pdf.html#:~:text=Choose%20%E2%80%9CTools%E2%80%9D%20%3E%20%E2%80%9COrganize,or%20pages%20from%20the%20file.}{Adobe Acrobat} (on all OSs), as well as \href{https://superuser.com/questions/517986/is-it-possible-to-delete-some-pages-of-a-pdf-document}{command line tools}.

\end{document}